\documentclass{article}
\usepackage{ijcai23}

\usepackage{times}
\usepackage{soul}
\usepackage{url}
\usepackage[hidelinks]{hyperref}
\usepackage[utf8]{inputenc}
\usepackage[small]{caption}
\usepackage{graphicx}
\usepackage{amsmath}
\usepackage{amsthm}
\usepackage{booktabs}
\usepackage{amssymb}
\usepackage[switch]{lineno}
\usepackage{multirow}
\usepackage{makecell}
\usepackage{threeparttable}
\usepackage{subfigure}
\usepackage{bm}
\usepackage[linesnumbered,ruled,lined]{algorithm2e}
\def\method{SGDA}

\title{Semi-supervised Domain Adaptation in Graph Transfer Learning}

\author{
Ziyue Qiao$^{1,2,3}$
\and
Xiao Luo$^{2}$\and
Meng Xiao$^{4,5}$\and
Hao Dong$^{4,5}$\and
Yuanchun Zhou$^{4,5}$\And
Hui Xiong$^{2,3,}$\footnotemark[2]
\affiliations
$^1$Jiangmen Laboratory of Carbon Science and Technology, Jiangmen\\
$^2$The Hong Kong University of Science and Technology (Guangzhou), Guangzhou\\
$^3$Guangzhou HKUST Fok Ying Tung Research Institute, Guangzhou\\
$^4$Computer Network Information Center, Chinese Academy of Sciences, Beijing\\
$^5$University of Chinese Academy of Sciences, Beijing
\emails
\{ziyuejoe, xiaoluopku, xiaomeng7890, donghcn\}@gmail.com,
zyc@cnic.cn,
xionghui@ust.hk
}

\begin{document}

\maketitle

\renewcommand{\thefootnote}{\fnsymbol{footnote}}
\footnotetext[2]{Corresponding author.}

\begin{abstract}
As a specific case of graph transfer learning, unsupervised domain adaptation on graphs aims for knowledge transfer from label-rich source graphs to unlabeled target graphs. However, graphs with topology and attributes usually have considerable cross-domain disparity and there are numerous real-world scenarios where merely a subset of nodes are labeled in the source graph. This imposes critical challenges on graph transfer learning due to serious domain shifts and label scarcity. 
To address these challenges, we propose a method named \textbf{S}emi-supervised \textbf{G}raph \textbf{D}omain \textbf{A}daptation (\method{}).
To deal with the domain shift, we add adaptive shift parameters to each of the source nodes, which are trained in an adversarial manner to align the cross-domain distributions of node embedding, thus the node classifier trained on labeled source nodes can be transferred to the target nodes. 
Moreover, to address the label scarcity, we propose pseudo-labeling on unlabeled nodes, which improves classification on the target graph via measuring the posterior influence of nodes based on their relative position to the class centroids. 
Finally, extensive experiments on a range of publicly accessible datasets validate the effectiveness of our proposed \method{} in different experimental settings.
\end{abstract}


\section{Introduction}

In the real world, graphs have been gaining popularity for their ability to represent structured data. 
As a basic problem, node classification has been applied in a variety of scenarios, including social networks~\cite{fan2019graph,ju2023comprehensive}, academic networks~\cite{kong2019academic,wu2020comprehensive}, and biological networks~\cite{ingraham2019generative,wang2018network}.
Graph transfer learning, which transfers knowledge from a labeled source graph to help predict the labels of nodes in a target graph with domain changes, has attracted a growing amount of interest recently. This problem is crucial due to the prevalence of unlabeled graphs in the real world and the anticipation of acquiring information from known domains.

Despite the significant progress made by graph transfer learning algorithms, they often assume that all nodes in the source graph are labeled. However, annotating the whole source graph becomes time-consuming and costly, particularly for large-scale networks. 
It is worth noting that recent semi-supervised node classification approaches can produce superior performance with a small number of node labels. This raises a natural problem,
whether it is possible to use a small number of labeled data and a large amount of unlabeled data in the source network to infer label semantic information in the target graph with significant domain discrepancy. In a nutshell, this innovative application scenario is summarized as a semi-supervised domain adaptation on graphs.

\begin{figure}
    \centering
\includegraphics[width=0.44\textwidth]{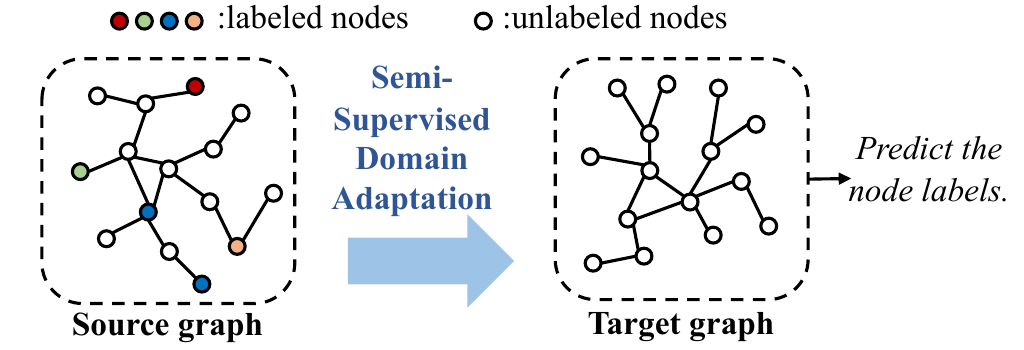}
    \caption{The semi-supervised domain adaptation on graphs.}
    \label{fig:intro}
\end{figure}

Nonetheless, formalizing a semi-supervised domain adaptive framework for node classification remains a non-trivial task since it must address two basic issues: 
\textbf{Issue 1: How to overcome a significant domain shift cross graphs to give domain-invariant predictions?} 
The domain shift between the source and target graphs roughly lies in the following two views: graph topology and node attributes. For example, distinct graphs may have different link densities and substructure schemas, and hand-crafted node attributes from diverse sources could have significant biases individually. This brings more considerable domain shifts than traditional data.
\textbf{Issue 2: How to mitigate the label scarcity for the classifier to give accurate and label-discriminative predictions?} 
Since the target graph is completely unlabeled, existing work only performs domain alignment, without considering the situation that the overall distributions may be aligned well but the class-level distributions may not match the classifier well.
Even worse, only a subset of node labels is available on the source graph, impeding effective classifier learning. Thus, it is critical to leverage the graph topology to improve the discriminative ability of the classifier on the unlabeled nodes.

In this paper, we address the aforementioned problems by developing a novel \textbf{S}emi-supervised \textbf{G}raph \textbf{D}omain \textbf{A}daptation model named \method{}, as shown in Figure \ref{fig:intro}.
The core idea of \method{} is to transfer the label knowledge from the source graph to the target graph via adversarial domain transformation and improve the model's prediction on unlabeled target nodes via adaptive pseudo-labeling with posterior scores.
Specifically, we add pointwise node mutual information into the graph encoder, enabling the exploration of high-order topological proximity to learn generalized node representations. 
In addition, we add shift parameters onto the source graph to align the distribution of cross-domain node embeddings, so as to the classifier trained on source node embedding can be used to predict the labels of the target node. We propose an adversarial domain transformation module to train the graph encoder and shift parameters. 
Furthermore, to address the label scarcity problem, we introduce pseudo labels to supervise the training of unlabeled nodes in both domains, which adaptively increases the training weights of nodes close to the pseudo labels' cluster centroids, thus making the model gives discriminative predictions on these unlabeled nodes.
The main contributions of our method for the semi-supervised domain adaptation in graph transfer learning can be summarized as follows:

\begin{itemize}
    \item To eliminate the domain shift cross graphs, we introduce the concept of shift parameters on the source graph encoding and propose an adversarial transformation module to learn domain-invariant node embeddings. 
    \item To alleviate the label scarcity, we propose a novel pseudo-labeling method using posterior scores to supervise the training of unlabeled nodes, improving the discriminative ability of the model on the target graph.
    \item Extensive experiments on various graph transfer learning benchmark datasets demonstrate the superiority of our \method{} over state-of-the-art methods.

\end{itemize}
\section{Related Works}

\noindent\textbf{Domain Adaptation.}
Domain adaptation aims to transfer semantic knowledge from a source domain to a target domain, which has various applications in computer vision~\cite{ijcai2022p232,ijcai2022p213}. In the literature, current methods can be roughly categorized into two types, i.e., distance-based methods~\cite{chang2021unsupervised,li2020enhanced,zhang2021deep} and adversarial learning-based methods~\cite{zhang2018collaborative,tzeng2017adversarial,volpi2018adversarial}. Distance-based methods explicitly calculate the distribution distance between source and target domains and minimize them in the embedding space. Typical metrics for distribution difference include maximum mean discrepancy (MMD)~\cite{chang2021unsupervised} and enhanced transport distance (ETD)~\cite{li2020enhanced}. 
Adversarial learning-based methods usually train a domain discriminator on top of the hidden embeddings and attempt to fuse it for domain alignment in an implicit fashion.
Despite the enormous effectiveness of domain adaptation, these methods typically focus on image problems. We address the challenge of semi-supervised domain adaptation on graphs by exploiting the graph topology information to enhance the model performance.

\noindent\textbf{Graph Transfer Learning.}
Graph transfer learning has been widely studied in recent years. Early models~\cite{qiao2022rpt,qiu2020gcc,hu2020gpt} typically utilize source data to construct a graph model for a different but related task in the target data. The effectiveness of transfer learning on graphs has been widely validated in a multi-task learning paradigm. 
Thus, graph transfer learning alleviates the burden of collecting labels regarding new tasks. The recent focus has been transformed into the problem of domain adaptation on graphs. 
Typically, these methods~\cite{guo2022learning,shen2020adversarial} combine graph model with domain adaption techniques. In particular, they produce domain-invariant node representations either implicitly confounding a domain discriminator using adversarial learning~\cite{zhang2021adversarial,wu2020unsupervised} or explicitly minimizing the distance~\cite{shen2020network} between representations in two domains.  
Still, most work establishes methods on graphs similar to those on images, without considering the complex structure of graphs or explicitly exploiting the graph topology information.

\noindent\textbf{Semi-supervised Learning on Graphs.}
Semi-supervised learning on graphs refers to the node classification problem, where only a small subset
of nodes are labeled.
Graph neural networks (GNNs) such as GCN~\cite{welling2016semi}, GraphSAGE~\cite{hamilton2017inductive}, and GAT~\cite{velivckovic2018graph} have achieved great success on these problem.
These methods usually follow the paradigm of message passing where each node attains information from its connected neighbors, followed by an aggregation operation for node representation updating in a recursive fashion. 
Recently, a range of GNN methods have been proposed to enhance the model performance from the view of exploring augmentation~\cite{wen2022adversarial,wang2020nodeaug}, expanding continuous~\cite{xhonneux2020continuous},  adversarial learning~\cite{xu2022unsupervised,jin2021adversarial} and etc~\cite{qiao2023dual,tang2021data}. However, these methods usually focus on learning and evaluation from a single graph. By contrast, we investigate a novel graph transfer learning problem named semi-supervised domain adaptation on graphs in this paper.

\begin{figure*}
    \centering
    \includegraphics[width=0.83\textwidth]{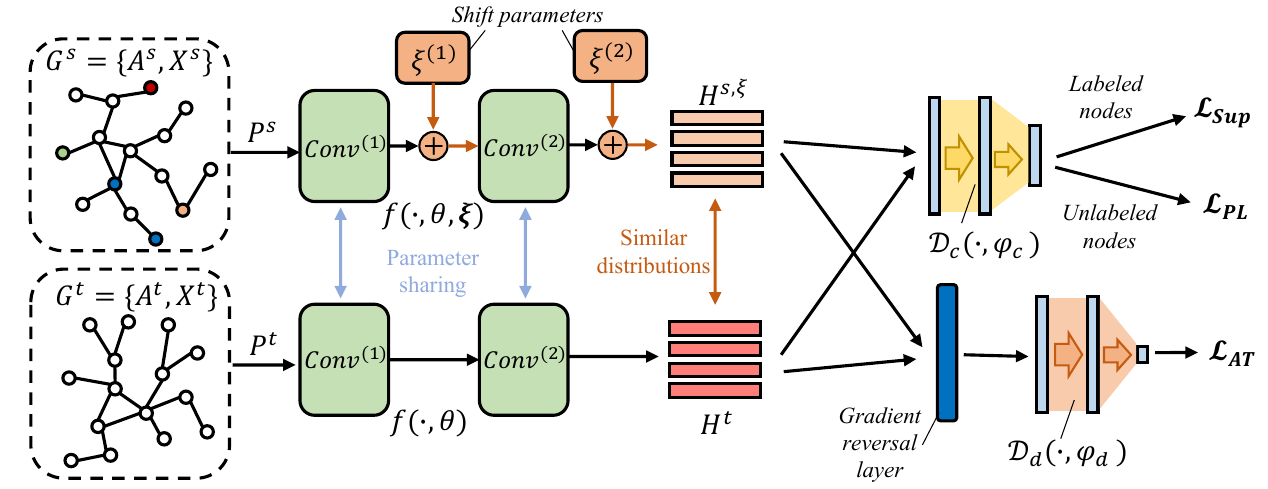}
    \caption{The framework of \method{}. The source graph and target graph with reconstructed high-order topologies $P^s$ and $P^t$ are fed into a two-layer graph convolutional network to generate generalized node embeddings, where source graph are added with shift parameters $\bm{\xi}$ to promote distribution alignment. Three losses $\mathcal{L}_{Sup}$, $\mathcal{L}_{AT}$, and $\mathcal{L}_{PL}$ perform supervised learning, domain adversarial transformation via shifting, and pseudo-labeling with posterior scores, respectively.
    }
    \label{fig:frame}
\end{figure*}

\section{Problem Definition}

The \textit{source graph} is expressed as $G^s = \{A^s, X^s\}$ with $\mathcal{V}^{s,l},\mathcal{V}^{s,u}$, and $Y^{s,l}$, where $A^s \in \mathbb{R}^{N^s\times N^s}$ is the adjacency matrix and $N^s$ is the number of nodes in $G^s$. $A^{s}_{ij} =1$ if there is an edge between nodes $n_i$ and $n_j$, otherwise, $A^{s}_{ij} =0$. $X^s\in \mathbb{R}^{N^s\times d}$ is the attribute matrix, where $d$ is the dimension of node attributes. $\mathcal{V}^{s,l}$ is the labeled node set, $y_i\in \mathbb{R}^{C}$ is the ground-truth of the node $n_i\in \mathcal{V}^{s,l}$ where $C$ is the number of classes and the $k$-th element $y_{i,k} = 1$ if the $i$-th node belong to the $k$-th class and otherwise $y_{i,k} = 0$. $\mathcal{V}^{s,u}$ is rest unlabeled node set in $G^{s}$, i.e, $|\mathcal{V}^{s,l}| + |\mathcal{V}^{s,u}| = N^{s}$.  $|\mathcal{V}^{s,l}|$ is much fewer than $|\mathcal{V}^{s,u}|$ due to the expensive labeling cost. 

The \textit{target graph} is expressed as $G^t = \{A^t, X^t\}$ with node set $\mathcal{V}^{t}$, where $A^t \in \mathbb{R}^{N^t\times N^t}$ is adjacency matrix and $N^t$ is number of nodes in $G^t$. $X^t\in \mathbb{R}^{N^t\times d}$ is the node attribute matrix. Note that the attribute set of the source and target graphs may have certain differences. However, one can create a union attribute set between them to align the dimension.

The problem of \textit{semi-supervised domain adaptation on graphs} is given the source graph $G^s$ with limited labels and the target graph $G^t$ completely unlabeled, and they have certain domain discrepancy on the data distributions but share the same label space. 
The goal is to learn a model to accurately predict the node classes in the target graph with the assistance of the partially labeled source graph. 

\section{Methodology}

As shown in Figure \ref{fig:frame}, our \method{} consists of three modules as below:
(1) \textit{Node Embedding Generalization.} To sufficiently explore high-order structured information in both graphs to learn generalized node representations;
(2) \textit{Adversarial Transformation.} To eliminate serious domain discrepancy between the source graph and the target graph, we introduce adaptive distribution shift parameters to the source graph, which are trained in an adversarial manner with regard to a domain discriminator. Therefore, the source graph is equipped with the target distribution.  
(3) \textit{Pseudo-Labeling with Posterior Scores.} To alleviate the label scarcity, we propose pseudo-labeling loss on all unlabeled nodes cross-domain, which improves classification on the target graph via measuring the influence of nodes adaptively based on their relative position to the class centroid.

\subsection{Node Embedding Generalization}
\label{sec:ppmi}
Considering that the model needs to perform the cross-domain transfer and the labels are limited for the classification task, learning generalized node embeddings is critical for such a domain adaptation procedure.
In view of this, 
we compute the positive pointwise mutual information~\cite{zhuang2018dual} between nodes to fully explore high-order unlabeled graph topology information and use the graph convolutional network~\cite{welling2016semi} to encode nodes into generalized low-dimensional embeddings.

Given a graph $G = \{A,X\}$ with the adjacency matrix $A\in \mathbb{R}^{N\times N}$, 
We use the random walk to sample a set of paths on $A$ and obtain a co-occurrence frequency matrix $F\in \mathbb{R}^{N\times N}$, where $F_{ij}$ counts the times of the node $n_j$ occurs within a predefined window in node $n_i$'s context. 
Then, the positive mutual information between nodes is computed by:

\begin{equation}
\begin{aligned}
    \mathbb{P}_{ij} = \frac{F_{ij}}{\sum_{i,j}F_{ij}}&, \text{ }
    \mathbb{P}_{i} = \frac{\sum_{j}F_{ij}}{\sum_{i,j}F_{ij}}, \text{ } 
    \mathbb{P}_{j} = \frac{\sum_{i}F_{ij}}{\sum_{i,j}F_{ij}}, \\
    P_{ij} &= max\{log(\frac{\mathbb{P}_{ij}}{\mathbb{P}_{i}\times \mathbb{P}_{j}}), 0\},
\end{aligned}
\end{equation}
where $\mathbb{P}_{ij}$ is the probability of the node $n_j$ occurring in the context of the node $n_i$.  $\mathbb{P}_{i}$ and $\mathbb{P}_{j}$ are the probability of the node $n_i$ as the anchor and node $n_j$ as the context, respectively. 
$P_{ij}$ is the positive mutual information between $n_i$ and $n_j$, which reflects the high-order topological proximity between nodes, as it assumes that if two nodes have high-frequency co-occurrence, $P_{ij}$ should be greater than if they are expected independent. i.e., $\mathbb{P}_{ij}>\mathbb{P}_{i}\times \mathbb{P}_{j}$.
We can obtain a mutual information matrix $P$ as the new adjacency matrix of $G$. Then, the $l$-th graph convolutional layer $Conv^{(l)}(\cdot)$ is defined as:

\begin{equation}
\label{eq:conv}
\begin{aligned}
H^{(l)} &= Conv^{(l)}(P, H^{(l-1)}), \\
&=\sigma( {D^{-\frac{1}{2}}\widetilde{P} D^{-\frac{1}{2}}H^{(l-1)}W^{(l)}}),
\end{aligned}
\end{equation}
where $\sigma(\cdot)$ denotes an activation function. $\widetilde{P} = P +I$ where $I$ is the identity matrix and
$D$ is the diagonal degree matrix of $P$ (i.e., $D_{ii} = \sum_{j} \widetilde{P}_{ij}$). $W^{(l)}$ is the $l$-th layer weight matrix. $H^{(l)}$ is the $l$-th layer hidden output and $H^{(0)} = X$. Finally, we can build the backbone of our method by stacking $L$ layers of graph convolutional networks in Equation \ref{eq:conv}, expressed as $f(G;\theta)$, where $\theta$ is the model parameters.

\subsection{Adversarial Transformation via Shifting}

Usually, the general learning objective of the domain adaptation is to train a feature encoder to eliminate the distribution discrepancy between the source domain and the target domain and then generate embeddings with similar distribution on both domains. Therefore, the classifier learned on the source domain can be adapted to the target domain.
Most methods~\cite{zhang2021adversarial,xiao2022domain} attempt to match embedding space distributions by optimizing the feature encoder itself. However, graphs with non-Euclidean topology usually have more considerable input disparity than traditional data. 
Only using the parameters in the encoder (e.g., GNNs) may be insufficient to shift the distributions finely.
Performing transition by adding trainable parameters (e.g., transport, perturbation) on input spaces has been proven to be effective in shifting one distribution to another one~\cite{jiang2020bidirectional,li2020enhanced}. Be aware that, we proposed an adversarial transformation module, which aims to add shift parameters on the source graph to modify its distribution and use adversarial learning to train both the graph encoder and shift parameters to align the cross-domain distributions.

Specifically, given the source graph $G^s = \{A^s, X^s\}$ and the target graph $G^t = \{A^t, X^t\}$, we first add the shift parameters $\bm{\xi}$ onto the source graph and obtain the shifted source node embeddings $H^{s,\xi} = f(G^s; \theta,\bm{\xi})$ with the distribution $\mathcal{H}^{s,\xi}$, where $H^{s,\xi}\in \mathbb{R}^{N^s\times h}$ and $h$ is the output dimension. Meanwhile the target node embeddings are obtained by $H^t = f(G^t; \theta)$ with the distribution $\mathcal{H}^{t}$, where $H^t \in \mathbb{R}^{N^t\times h}$. 
The optimization objective is to make the distributions similar, i.e., $\mathcal{H}^{s,\xi} \approx \mathcal{H}^{t}$. 
We define the shift parameters as randomly initialized multi-layer parameter matrices, i.e., $\bm{\xi} = \{\xi^{(1)}, \xi^{(2)}, ..., \xi^{(L)}\}$, where each $\xi^{(i)}$ is specific to the $i$-th layer hidden output of $f(\cdot)$, formulated as:

\begin{equation}
H^{s,(l)} = 
\begin{cases}
Conv^{(l)}(P^s, X^s)+\mathbf{\xi}^{(l)} & l=1 \\
Conv^{(l)}(P^s, H^{s,(l-1)})+\mathbf{\xi}^{(l)} \quad & 1<l\leq L \\
\end{cases}
\end{equation}

Then, we can obtain the shifted source node embeddings $H^{s,\xi}$ from the final output. 
We propose an adversarial transformation optimization objective on the source node embeddings and target node embeddings, which is defined as:

\begin{equation}
\begin{aligned}
     \max_{\theta, \bm{\xi}}& \left\{  \min_{\phi_d}\left\{\mathcal{L}_{AT}(H^{s,\xi}, H^t; \phi_d)\right\}\right\}, \\
    & s.t., {||\xi^{(l)}||}_F\leq \epsilon, \forall \xi^{(l)}\in \bm{\xi}.
\end{aligned} 
\end{equation}

The loss function $\mathcal{L}_{AT}$ is defined as:

\begin{equation}
\label{eq:at}
\begin{aligned}
 \mathcal{L}_{AT} = &- \mathbb{E}_{h_i^{s,\xi}\sim \mathcal{H}^{s,\xi}}\left[log\left(\mathcal{D}_d(h^{s,\xi}_{i}, \phi_d)\right)\right]\\
&- \mathbb{E}_{h_j^{t}\sim \mathcal{H}^{t}}\left[log\left(1-\mathcal{D}_d(h^{t}_{j}, \phi_d)\right)\right],
\end{aligned}
\end{equation}
where $h^{s,\xi}_{i}, h^{t}_{i}$ is the $i$-th row of $H^{s,\xi}, H^t$, respectively.
$\mathcal{D}_d(h_{i}, \phi_d)$ with the parameters $\phi_d$ is a domain discriminator that learns a logistic regressor: $\mathcal{D}_d: \mathbb{R}^h \rightarrow \mathbb{R}^1$ to model the probability of the given the input node embedding $h_{i}$ from the source graph or the target graph. 
The domain discriminator is trained to distinguish which domain the node embeddings are from, while the encoder with shift parameters is forced to generate the source node embeddings as indistinguishable as possible from target ones for the domain discriminator, thus resulting in domain-invariant node embeddings,
We constrain the gradient of shift parameters in each training step within a certain radius $\epsilon$ to avoid excessive distribution shift, making the adversarial task impossible.

\subsection{Pseudo-Labeling with Posterior Scores}
In this module, we define the classifier on the node embeddings, e.g., $\mathcal{D}_c(h_i, \phi_c): \mathbb{R}^h \rightarrow \mathbb{R}^C$ to model the label probability of nodes, which is a multi-layer perception followed with a softmax layer. Then, we can obtain the probability $p_i^s = \mathcal{D}_c(h^{s,\xi}_{i}, \phi_c)$ of each node in the source graph and the probability $p_j^t = \mathcal{D}_c(h^{t}_{j}, \phi_c)$  of each node in the target graph.
We define the supervised loss function on the probabilities of nodes in the labeled node set $\mathcal{V}^{s,l}$ of the source graph:

\begin{equation}
\label{sup}
\mathcal{L}_{Sup} = -\frac{1}{|\mathcal{V}^{s,l}|} \sum_{n_{i} \in \mathcal{V}^{s,l}}\sum_{k=1}^C{y_{i,k}}\log(p_{i,k}^s).
\end{equation}

Since only a few nodes are labeled in the source graph whereas all nodes are unlabeled in the target graph, the model will be easily over-fitting if we only have Eq. \ref{sup}. In particular, without any supervision, the nodes in the target graph distributed near the border and far away from the centroid of clusters of their corresponding classes are easily misclassified by the hyperplane learned from the label information of the source graph. 
Thus, we propose a novel pseudo-labeling strategy with posterior scores of nodes to improve the prediction accuracy on unlabeled nodes.

Specifically, in each training iteration, we update the pseudo-labels for the unlabeled nodes in both the source and target graph by mapping their output probabilities into one-hot encoding, denoted as $\widehat{y}_{i} = \mathcal{M}(p_i)$, where $\mathcal{M}(\cdot)$ is the one-hot map and $p_{i}$ is the probability of node $n_i \in \mathcal{V}^{s,u}\cup\mathcal{V}^{t}$. 
Note that we treat all unlabeled nodes cross-domain in the same level and omit the notations of domain superscript for brevity.
We assume that the nodes close to the structural centroid of their pseudo-label cluster on the graph are more likely classified correctly, while the pseudo-labels of those close to the cluster boundary are less reliable. 
Based on the hypothesis, we treat the pseudo-labels of former nodes as more high-quality self-supervised signals and aim to improve the discriminative ability of these node embeddings.
Thus, we introduce a posterior score to define how $n_i$ is close to the structural centroid of its pseudo label cluster on its reconstructed adjacency matrix $P$ computed in Section \ref{sec:ppmi}:
\begin{equation}
    w_i = \sum_{j=1}^N (P_{ij} * P_{\mathcal{C}_{\widehat{y}_{i}},j} -  \frac{1}{C-1} \sum_{k=1,k\neq \widehat{y}_{i}}^C P_{ij} * P_{\mathcal{C}_{k},j}),
\end{equation}
where $P_{\mathcal{C}_{\widehat{y}_{i}},j} = \frac{1}{|\mathcal{C}_{\widehat{y}_{i}}|}\sum_{x\in \mathcal{C}_{\widehat{y}_{i}}} P_{x,j}$ indicates the overall mutual information from the nodes belonging to the class $\widehat{y}_{i}$ to the node $n_j$.
The posterior score defines that if a node $n_i$ with the pseudo label $\widehat{y}_{i}$ encounters high mutual information from other nodes in terms of the class $\widehat{y}_{i}$ and low ones from other nodes in terms of other classes, we have the conclusion that node $n_i$ close to the centroid of class $\widehat{y}_{i}$ and $w_i$ has a high value, and vice versa. 
Then, we apply a cosine annealing function~\cite{chen2021topology} to scale $w_i$ into a certain range:
\begin{equation}
    \widehat{w}_i = \alpha + \frac{1}{2}(\beta-\alpha)(1+\cos(\frac{Rank(w_i)}{|\mathcal{V}|}\pi)),
\end{equation}
where $\mathcal{V} \in \{\mathcal{V}^{s,u}, \mathcal{V}^{t}\}$ and $n_i \in\mathcal{V}$. $[\alpha, \beta]$ controls the scale range. $Rank(w_i)$ is the ranking order of $w_i$ from the largest to the smallest. 
Then, we define a pseudo-labeling loss function with posterior scores as follows:
\begin{equation}
    \mathcal{L}_{PL} =  -\frac{1}{|\mathcal{V}|} \sum_{n_{i} \in \mathcal{V}}\widehat{w}_i \sum_{k=1}^C{\widehat{y}_{i,k}}\log(p_{i,k}) + \sum_{k=1}^C \widehat{p}_k \log\widehat{p}_k,
\end{equation}
where $\widehat{p}_k = \mathbb{E}_{n_i\in  \mathcal{V}}[p_{i,k}]$ and the second term is a diversity regularization to promote the diversity of output probabilities, which can circumvent the problem of some large posterior scores dominating in training to make all unlabeled nodes over-fit into the same pseudo-label. By $\mathcal{L}_{SL}$, the model is more encouraged to focus on the high-confidence nodes close to its corresponding cluster centroid and less influenced by those ambiguous nodes near the boundary, so as to improve the discriminative ability on unlabeled nodes.

\subsection{Optimization}

By combining the three losses above, the optimization of the proposed method \method{} is as follows:

\begin{equation}
\begin{aligned}
    \min_{\theta, \bm{\xi}, \phi_c}& \left\{ \mathcal{L}_{Sup} + \lambda_1 \mathcal{L}_{PL} + \lambda_2 \max_{\phi_d} \{-\mathcal{L}_{AT}\}\right\}, \\
    & s.t., {||\xi^{(l)}||}_F\leq \epsilon, \forall \xi^{(l)}\in \bm{\xi}.
\end{aligned}
\end{equation}
where $\lambda_1$ and $\lambda_2$ is the weights to balance different losses. 
In practice, we introduce a gradient reversal layer (GRL)~\cite{ganin2016domain} between the graph encoder and the domain discriminator so as to conveniently perform min-max optimization under $\mathcal{L}_{AT}$ in one training step. The GRL acts as an identity transformation during the forward propagation and changes the signs of the gradient from the subsequent networks during the backpropagation.
Particularly, for each $\xi^{(l)}\in \bm{\xi}$, the update rule is defined as follow:

\begin{equation}
\begin{aligned}
& g(\xi^{(l)}) = \left(\frac{\partial\mathcal{L}_{Sup}}{\partial\xi^{(l)}} + \lambda_1 \frac{\partial\mathcal{L}_{PL}}{\partial\xi^{(l)}}- \lambda_2 \frac{\partial\mathcal{L}_{AT}}{\partial\xi^{(l)}}\right),\\
& \xi^{(l)} \leftarrow  \xi^{(l)} + \mu g(\xi^{(l)})/||g(\xi^{(l)})||_{F} .
\end{aligned}
\end{equation}
where $\mu$ is the learning rate. The shift parameters $\bm{\xi}$ are optimized by Projected Gradient Descent (PGD). 
Following \cite{yang2021graph,kong2020flag},  we use the unbounded adversarial transformation as one is not aware of the shifting scale in advance.

\section{Experiments}
\subsection{Dataset}
We conduct experiments on three real-world graphs: \textit{ACMv9} (A), \textit{Citationv1} (C), and \textit{DBLPv7} (D) from ArnetMiner~\cite{tang2008arnetminer}.
These graphs are constructed from three different source datasets in different periods, i.e., Association for Computer Machinery (after the year 2010), Microsoft Academic Graph (before the year 2008), and DBLP Computer Science Bibliography (between years 2004 and 2008), respectively so that they have varied distributions in their domain spaces. 
Each node in these graphs represents a paper whose attribute is the sparse bag-of-words vector of the paper's title. 
The edges represent a citation relationship between these papers, where the direction is ignored. 
As these graphs do not share the same feature set of node attributes, we union their attribute set and reshape the attribute dimension as 6775. 
Each node is assigned a five-classes label based on its relevant research areas, including \textit{Artificial Intelligence}, \textit{Computer Version}, \textit{Database}, \textit{Information Security}, and \textit{Networking}. 
Table~\ref{tab:dataset} presents the statistics of graph scale, attributes, average degree, and label proportion, indicating the intrinsic discrepancy between the three graphs.
In our paper, we alternately select one of these graphs as the source domain and the rest two as the target domain. 

\begin{table}[t]
    \centering
    \resizebox{\linewidth}{!}{
    \begin{tabular}{c|ccccc}
        \toprule
        Dataset & \#Nodes & \#Edges & \#Attr. & \makecell[c]{Avg.\\Degree}& \makecell[c]{Label Proportion\\ (\%)} \\ 
        \midrule
        ACMv9 & 9,360 & 15,602 & 5,571 & 1.667 & 20.5/29.6/22.5/8.6/18.8 \\
        Citationv1 & 8,935 & 15,113 & 5,379 & 1.691 & 25.3/26.0/22.5/7.7/18.5\\
        DBLPv7 & 5,484 & 8,130 & 4,412 & 1.482 & 21.7/33.0/23.8/6.0/15.5 \\ 
        \bottomrule
    \end{tabular}}
    \caption{The statistics of three graphs. `\#' means `the number of'. `Attr.' means `Attributes'. `Avg.' means `Average'.}
    \label{tab:dataset}
\end{table}

\subsection{Baselines}
We select two groups of baseline methods. The first group is traditional solutions for graph semi-supervised learning, which learns a node classification model on the source graph and directly uses the model to perform inductive prediction on the target graph without explicit transfer learning. We first use Multi-Layer Perceptron (\textit{MLP}), which is directly trained on the attributes of nodes in the source graph. We choose four GNN variants, including GCN~\cite{welling2016semi}, GraphSAGE (\textit{GSAGE})~\cite{hamilton2017inductive}, GAT~\cite{velivckovic2018graph}, and GIN~\cite{xu2018powerful}, which are acknowledged as state-of-the-art models for graph semi-supervised learning. 
The second group is specific to domain adaptation. We first choose two general approaches DANN~\cite{ganin2016domain} and CDAN~\cite{ganin2016domain}, which are initially designed for transfer learning on images or text. We train them on node attributes. For graph semi-supervised learning, we create their variants DANN$_{GCN}$ and CDAN$_{GCN}$ by replacing the encoder from MLP to GCN and training them on graphs. Finally, we select two methods most similar to us, UDA-GCN~\cite{wu2020unsupervised} and AdaGCN~\cite{dai2022graph}, which are also designed for semi-supervised domain adaptation on graphs.

\begin{table*}[!htbp]
\centering

\resizebox{\linewidth}{!}{
\begin{tabular}{l|cccccccccccc}
\toprule
\multirow{2}{*}{Methods} & \multicolumn{2}{c}{A$\Rightarrow$C} & \multicolumn{2}{c}{A$\Rightarrow$D} & \multicolumn{2}{c}{C$\Rightarrow$A} & \multicolumn{2}{c}{C$\Rightarrow$D} & \multicolumn{2}{c}{D$\Rightarrow$A}  & \multicolumn{2}{c}{D$\Rightarrow$C} \\
 & Micro & Macro & Micro & Macro & Micro & Macro & Micro & Macro & Micro & Macro & Micro & Macro \\ \midrule
MLP & 41.3$_{\pm1.15}$ & 35.8$_{\pm0.72}$ & 42.8$_{\pm0.88}$ & 36.3$_{\pm0.77}$ & 39.4$_{\pm0.57}$ & 33.7$_{\pm0.58}$ & 43.7$_{\pm0.69}$ & 36.7$_{\pm0.55}$ & 37.3$_{\pm0.32}$ & 30.8$_{\pm0.37}$ & 39.4$_{\pm0.99}$ & 32.8$_{\pm0.99}$\\
GCN & 54.4$_{\pm1.52}$ & 52.0$_{\pm1.62}$ & 56.9$_{\pm2.33}$ & 53.4$_{\pm2.81}$ & 54.1$_{\pm1.40}$ & 52.3$_{\pm1.98}$ & 58.9$_{\pm0.99}$ & 54.5$_{\pm1.55}$ & 50.1$_{\pm2.14}$ & 48.0$_{\pm3.28}$ & 56.0$_{\pm1.24}$ & 51.9$_{\pm1.49}$\\
GSAGE & 49.3$_{\pm2.18}$ & 46.4$_{\pm2.06}$ & 51.8$_{\pm1.35}$ & 47.4$_{\pm1.62}$ & 46.8$_{\pm2.56}$ & 45.0$_{\pm2.78}$ & 51.7$_{\pm1.95}$ & 48.1$_{\pm1.97}$ & 41.7$_{\pm2.17}$ & 37.4$_{\pm4.59}$ & 45.4$_{\pm2.11}$ & 39.3$_{\pm3.45}$ \\
GAT  & 55.1$_{\pm3.22}$ & 50.8$_{\pm1.45}$ & 55.3$_{\pm2.52}$ & 51.8$_{\pm2.60}$ & 50.0$_{\pm1.20}$ & 45.6$_{\pm2.36}$ & 55.4$_{\pm2.73}$ & 49.2$_{\pm2.59}$ & 44.8$_{\pm2.74}$ & 38.3$_{\pm4.84}$ & 50.4$_{\pm3.35}$ & 42.0$_{\pm4.46}$\\
GIN & 64.6$_{\pm2.47}$ & 56.0$_{\pm2.73}$ & 60.0$_{\pm2.09}$ & 51.3$_{\pm3.99}$ & 57.1$_{\pm1.19}$ & 54.4$_{\pm2.57}$ & 62.0$_{\pm1.05}$ & 56.8$_{\pm1.40}$ & 51.9$_{\pm2.00}$ & 45.4$_{\pm2.16}$ & 60.2$_{\pm3.05}$ & 53.0$_{\pm2.10}$\\ \midrule
DANN & 44.3$_{\pm2.03}$ & 39.3$_{\pm1.86}$ & 44.0$_{\pm1.42}$ & 38.7$_{\pm1.47}$ & 41.8$_{\pm1.95}$ & 37.6$_{\pm1.24}$ & 45.5$_{\pm0.71}$ & 39.6$_{\pm1.55}$ & 37.8$_{\pm3.66}$ & 33.2$_{\pm2.23}$ &41.7$_{\pm2.32}$ & 35.6$_{\pm2.55}$\\
CDAN & 44.6$_{\pm1.30}$  & 38.6$_{\pm1.07}$  & 45.5$_{\pm0.85}$ & 38.0$_{\pm0.86}$ & 42.4$_{\pm0.64}$ & 36.2$_{\pm1.17}$ & 46.7$_{\pm1.17}$ & 39.2$_{\pm0.96}$ & 39.0$_{\pm1.08}$ & 32.3$_{\pm1.09}$ & 41.7$_{\pm1.55}$ & 34.8$_{\pm1.56}$ \\
DANN$_{GCN}$ & 63.0$_{\pm6.75}$ & 59.6$_{\pm6.02}$ & 62.2$_{\pm1.90}$ & 57.7$_{\pm3.16}$ & 56.7$_{\pm0.38}$ & 55.2$_{\pm1.03}$ & 65.3$_{\pm2.04}$ & 59.0$_{\pm2.39}$ & 52.3$_{\pm2.59}$ & 48.6$_{\pm4.52}$ &58.1$_{\pm2.78}$ & 52.4$_{\pm3.81}$ \\
CDAN$_{GCN}$ &70.3$_{\pm0.84}$ & 66.5$_{\pm0.66}$ & 65.0$_{\pm1.00}$ & 61.3$_{\pm0.96}$ & 56.3$_{\pm1.78}$ & 53.6$_{\pm2.70}$ & 65.2$_{\pm2.19}$ & 58.8$_{\pm2.38}$ & 53.0$_{\pm1.34}$ & 48.7$_{\pm3.51}$ & 59.0$_{\pm1.52}$ & 53.3$_{\pm1.99}$ \\
UDA-GCN & 72.4$_{\pm2.75}$ & 65.2$_{\pm6.51}$ & 68.0$_{\pm6.38}$ & 64.3$_{\pm7.12}$ & 62.9$_{\pm0.33}$ & 62.2$_{\pm1.44}$ & 71.4$_{\pm2.56}$ & 67.5$_{\pm2.25}$ & 55.8$_{\pm3.50}$ & 52.4$_{\pm2.68}$ & 65.2$_{\pm4.41}$ & 60.7$_{\pm6.84}$\\
AdaGCN & 70.8$_{\pm0.95}$ & 68.5$_{\pm0.73}$ & 68.2$_{\pm3.84}$ & 64.2$_{\pm3.91}$ & 61.5$_{\pm2.20}$ & 60.4$_{\pm3.15}$ & 69.1$_{\pm1.96}$ & 65.8$_{\pm2.87}$ & 56.1$_{\pm1.75}$ & 53.8$_{\pm2.95}$ & 64.1$_{\pm0.91}$ & 62.8$_{\pm1.56}$ \\ 
\midrule

\method{} &\textbf{75.6}$\mathbf{_{\pm0.57}}$ & \textbf{71.4}$_{\mathbf{\pm0.82}}$& \textbf{69.2}$\mathbf{_{\pm0.73}}$ & \textbf{64.7}$\mathbf{_{\pm2.36}}$ & \textbf{66.3}$\mathbf{_{\pm0.68}}$ & \textbf{62.3}$\mathbf{_{\pm0.96}}$ & \textbf{72.9}$\mathbf{_{\pm1.26}}$ & \textbf{68.9}$\mathbf{_{\pm1.83}}$ & \textbf{60.6}$\mathbf{_{\pm0.86}}$ & \textbf{56.0}$\mathbf{_{\pm0.90}}$ & \textbf{73.2}$\mathbf{_{\pm0.59}}$ & \textbf{69.3}$\mathbf{_{\pm1.01}}$ \\

\bottomrule
\end{tabular}}

\caption{The model performance comparison on six domain adaptation tasks with source label rate as 5\%. A: ACMv9; C:Citationv1; D: DBLPv7. A$\Rightarrow$C represents that A is the source graph and C is the target graph. The same applies to other tasks.}
\label{tab:results}
\end{table*}

\subsection{Experimental Setting}
We choose a two-layer GCN as the backbone of \method{}. We set the loss wrights $\lambda_1$ always as 1 and $\lambda_2 \in [0,1]$ as a dynamic value that is linearly increased with the training epoch, i.e., $\lambda_2 = m/M$ where $m$ is the current epoch and $M$ is the maximum epoch. We consider that at the early training steps, the classifier is not completely converged, making the pseudo labels produced by the classifier inferior for self-supervised learning. 
We randomly initialize the $\bm{\xi}$ under the uniform distribution $U(-\epsilon,\epsilon)$ and set the $\epsilon$ always as $0.5$. 
We set the scale range $\alpha$ and $\beta$ always as 0.8 and 1.2.
We train \method{} for 200 epochs with the learning rate as $0.001$, the weight decay as $0.001$, and the dropout rate as $0.1$ on all datasets.
For baselines, we implement DANN, CDAN, AdaGCN, and UDA-GCN by their own codes and report the best results. 
The dimension of node embeddings is as 512 for all approaches.

\subsection{Performance Comparison}
This experiment aims to answer: \textit{How is \method{}{} performance on the semi-supervised domain adaptation task on graphs?} 
We randomly select 5\% of nodes in the source graph as labeled nodes and others as unlabeled nodes while the target graph is completely unlabeled. We use Micro-F1 and Macro-F1 as the metric and report the classification results of different approaches on the target graph in Table \ref{tab:results}. 
Notably, we ran each experiment 5 times, and each time we sampled different label sets to alleviate the randomness. The average results with standard deviation are reported. 

From the results, we can observe that:
Firstly, \method{} consistently achieved the best results on six transfer learning tasks. Particularly, \method{} achieved a significant improvement compared with two methods of semi-supervised domain adaption across graphs---AdaGCN and UDA-GCN, indicating the effectiveness of \method{} in solving this problem. 
Another observation is that the performance of MLP is worse than GNNs; two general domain adaptation methods DANN and CDAN are worse than their two variants DANN$_{GCN}$ and CDAN$_{GCN}$, indicating incorporating graph topology information is critical.
Thirdly, the domain adaptation methods generally performed better than the inductive learning methods. 
That proves that it is necessary to perform domain transformation to eliminate the cross-domain distribution discrepancy in the graph transfer learning task.
Lastly, as the domain adaptation methods with the GNN-based encoders, AdaGCN and UDA-GCN achieved better results than DANN$_{GCN}$ and CDAN$_{GCN}$. That is because they use more complex graph encoders and improved optimization objectives. Still, they are worse than \method{} due to the shortages of domain transformation and incompetence in handling the label scarcity.

\subsection{Ablation Study}

\begin{figure}[t]
\centering
\subfigure{
\includegraphics[width=4cm]{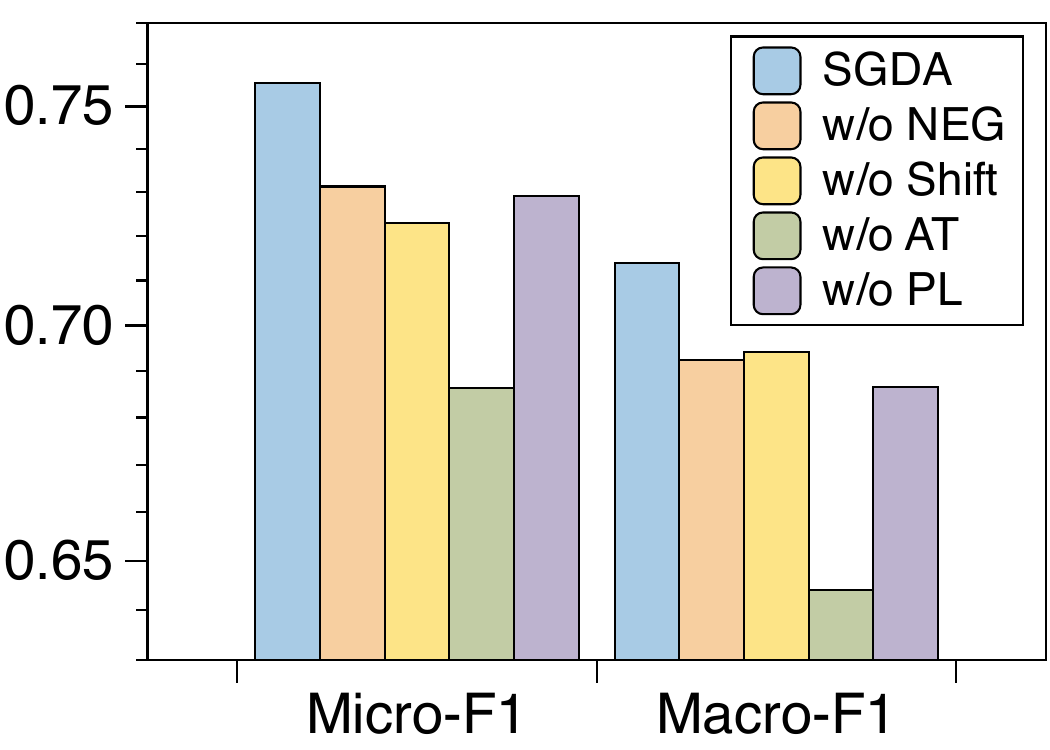}
}
\subfigure{
\includegraphics[width=4cm]{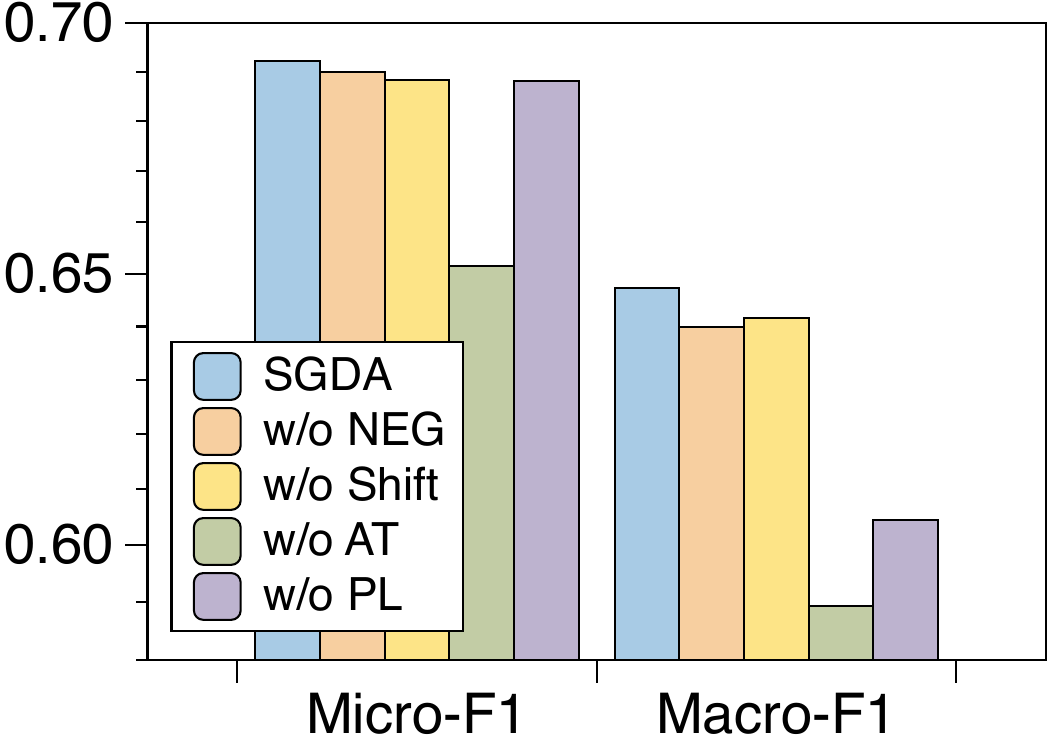}
}
\caption{The results of ablation study on the A $\Rightarrow$ C task (left) and the A $\Rightarrow$ D task (right). }
    \label{fig:ablation}

\end{figure}

This experiment aims to answer: \textit{Are all the proposed technologies of \method{} have the claimed contribution to the semi-supervised domain adaptation on graphs?}
For that, we design four variant methods for \method{} to verify the effectiveness of node embedding generalization, adversarial transformation via shifting, and pseudo-labeling: 
\textit{\textbf{w/o NEG}}: we directly use the original graph adjacency matrix rather than reconstruct the adjacency.
\textit{\textbf{w/o Shift}}: we remove the shift parameters added on the source graph and only use the graph encoder to learn domain-invariant node embeddings.
\textit{\textbf{w/o AT}}: we remove the loss $\mathcal{L}_{AT}$ so the  transformation of the source graph to the target graph is deactivated.
\textit{\textbf{w/o PL}}: we remove the loss $\mathcal{L}_{PL}$ so pseudo-labeling on unlabeled nodes is deactivated.
Figure~\ref{fig:ablation} reported the performance of these variants. 

\noindent\textbf{Effect of node embedding generalization.}\textit{w/o NEG} performs worse than \method{}. The reason is that without exploiting high-order graph topology information, the node embeddings only incorporated local neighborhood information are not generalized enough to perform the transformation.

\noindent\textbf{Effect of shift parameters.} \textit{w/o Shift} performs worse than \method{}. The reason is that the shift parameters in $\mathcal{L}_{AT}$ can facilitate the transfer ability of the model. Only utilizing the graph encoder is inefficient to shift distributions finely. 
Apart from this, we can observe that \textit{w/o Shift} is still superior to other SOTA domain adaptation methods. This phenomenon shows the limitation of those baselines on an incomplete labeled source graph and proves the advantage of $\mathcal{L}_{PL}$.

\noindent\textbf{Effect of adversarial transformation.} \textit{w/o AT} without any cross-domain distribution alignment performs worse but still achieves considerable performance compared with baselines in Table \ref{tab:results}. That is because our proposed pseudo-labeling on the target graph can adaptively pay more attention to those nodes close to the class centroid, who are more convinced to be correctly classified, and thus can constantly optimize the classifier to have better discrimination on the target graph.

\noindent\textbf{Effect of pseudo-labeling.} \textit{w/o PL} performs worse because, without $\mathcal{L}_{PL}$, the model is easier to get over-fitting on the source graph under limited labels. Additionally, the model may learn similar distributions between the source and target graph, but their class-level distributions may be inconsistent. The proposed pseudo-labeling can utilize the unlabeled graph topology to find better label space for nodes adaptively.

\begin{figure}[]
\centering
\subfigure[Source graph, UDAGCN]{
\includegraphics[width=3.9cm]{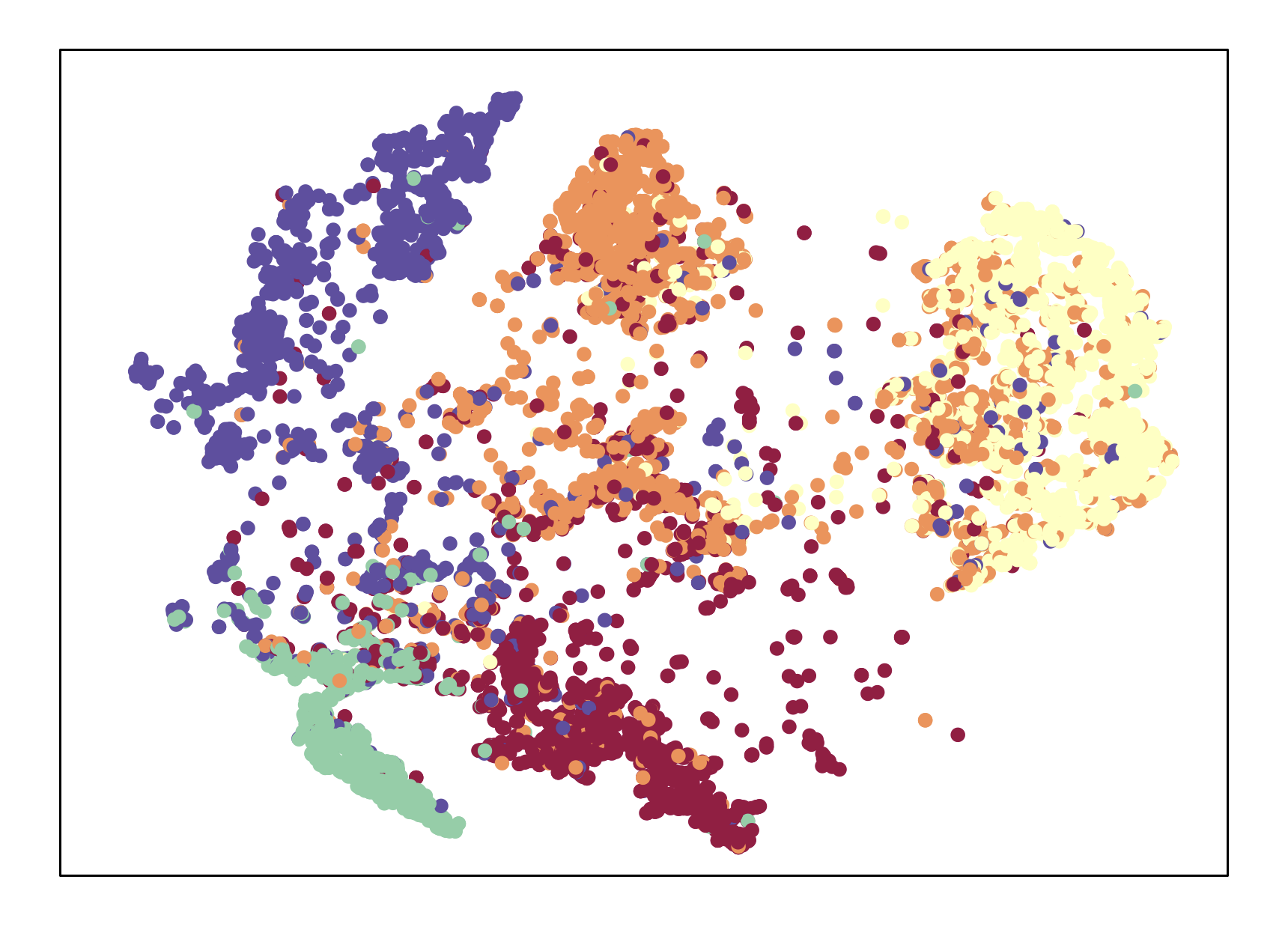}
}
\subfigure[Target graph, UDAGCN]{
\includegraphics[width=3.9cm]{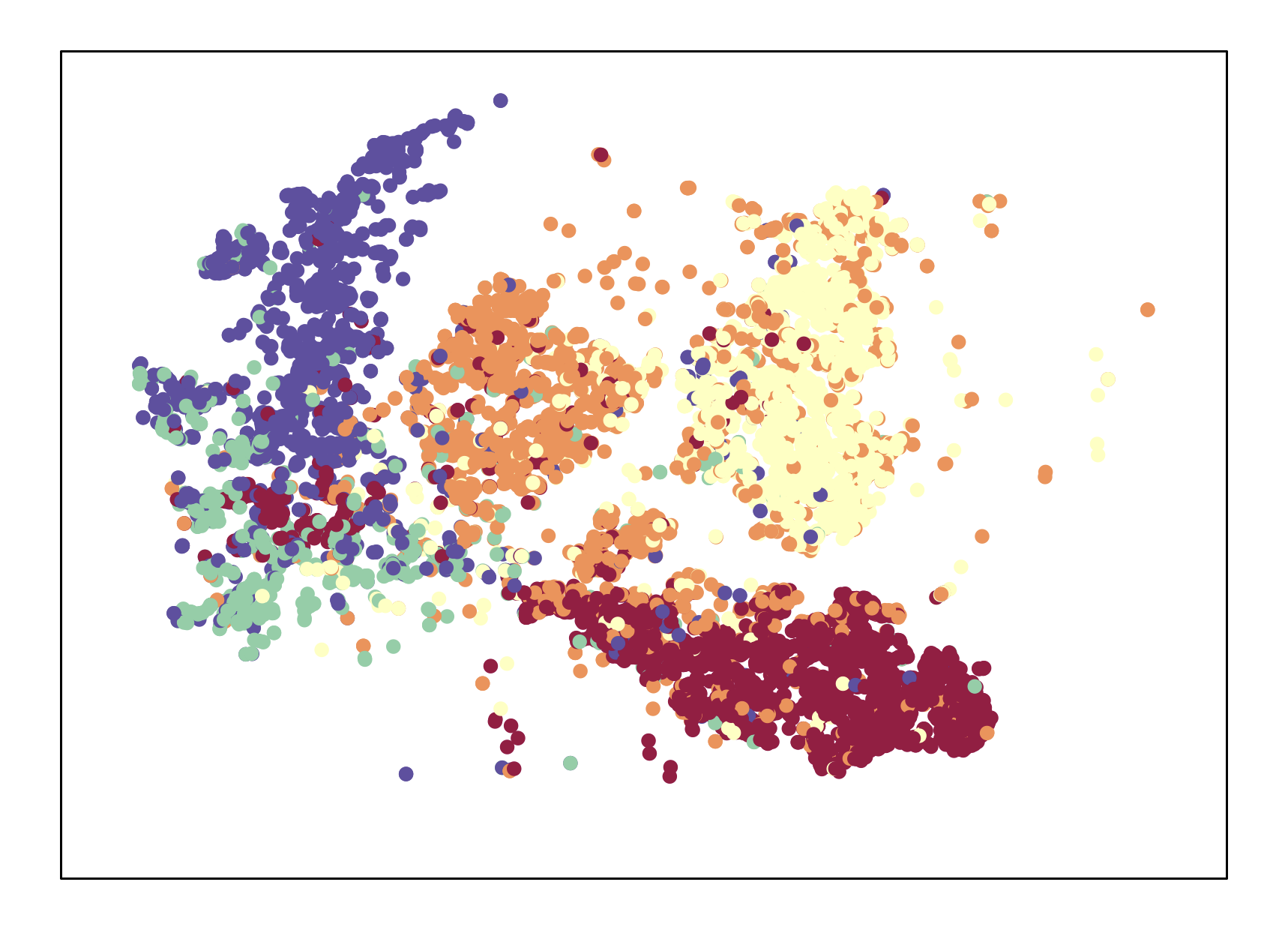}
}
\\ 
\subfigure[Source graph, \method{}]{
\includegraphics[width=3.9cm]{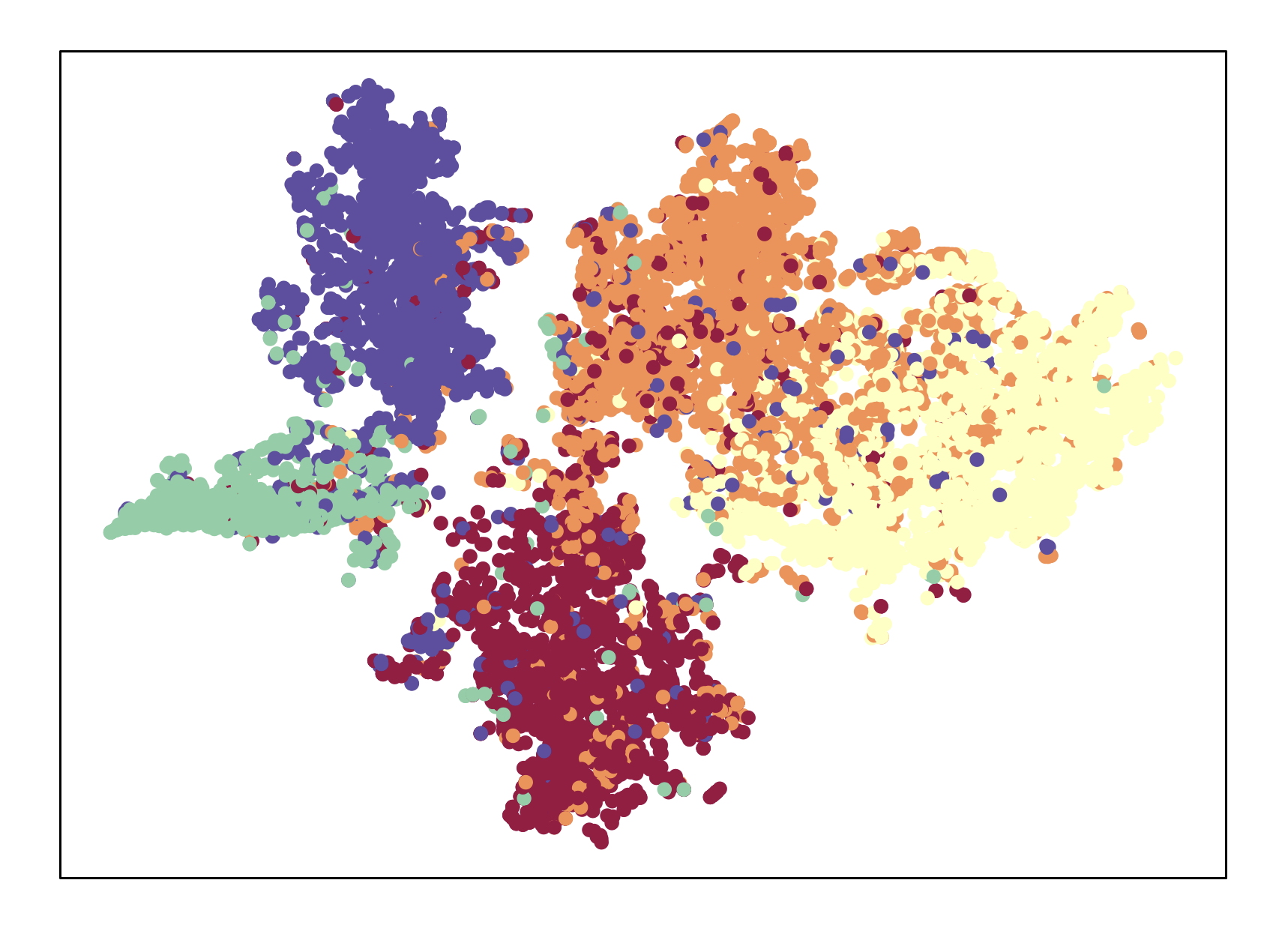}
}
\subfigure[Target graph, \method{}]{
\includegraphics[width=3.9cm]{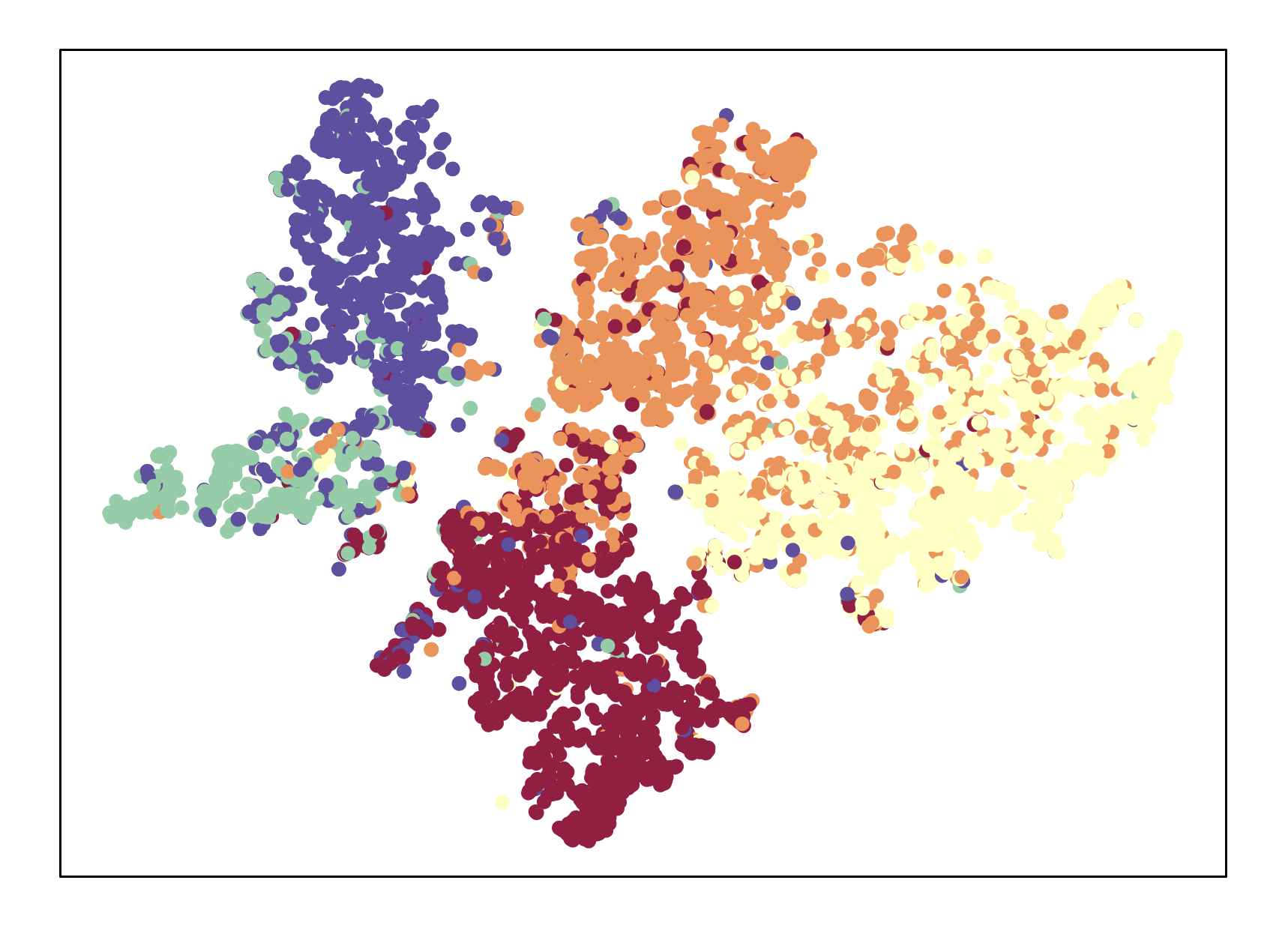}
}
\caption{Visualization of node embeddings learned by UDAGCN and \method{} on the A $\Rightarrow$ C task.}
\label{fig:visual}

\end{figure}

\subsection{Visualization of Distributions}
This experiment aims to answer that: \textit{How is the distribution of \method{}-generated node embeddings compared to the SOTA domain adaptation method?}
To illustrate the difference, we obtained the source and target node embeddings learned from \method{} and UDA-GCN with the same 5\% label rate setting. Then we separately projected them in 2-D by t-SNE and visualized them in Figure~\ref{fig:visual}. We colored each node by its class.
The first observation is that the distribution learned by \method{} could generate more generalized node embeddings, showing that nodes are dispersedly distributed in the space, which is contributed by preserving high-order topology information via random walks.
Also, the distributions of source space and target space learned by \method{} are clearly more consistent with each other, proving it can well eliminate the cross-domain discrepancy and learn domain-invariant node embeddings.
 Lastly, \method{} can significantly separate each class of nodes in both the source and the target graph. 
On the contrary, UDA-GCN can hardly differentiate each group of nodes, which is more apparent in the target space. 
That is because \method{} can well handle label scarcity and train a more discriminative classifier on the target nodes.

\begin{figure}[!ht]
\centering
\hspace{-4mm}
\subfigure{
\includegraphics[width=4.2cm]{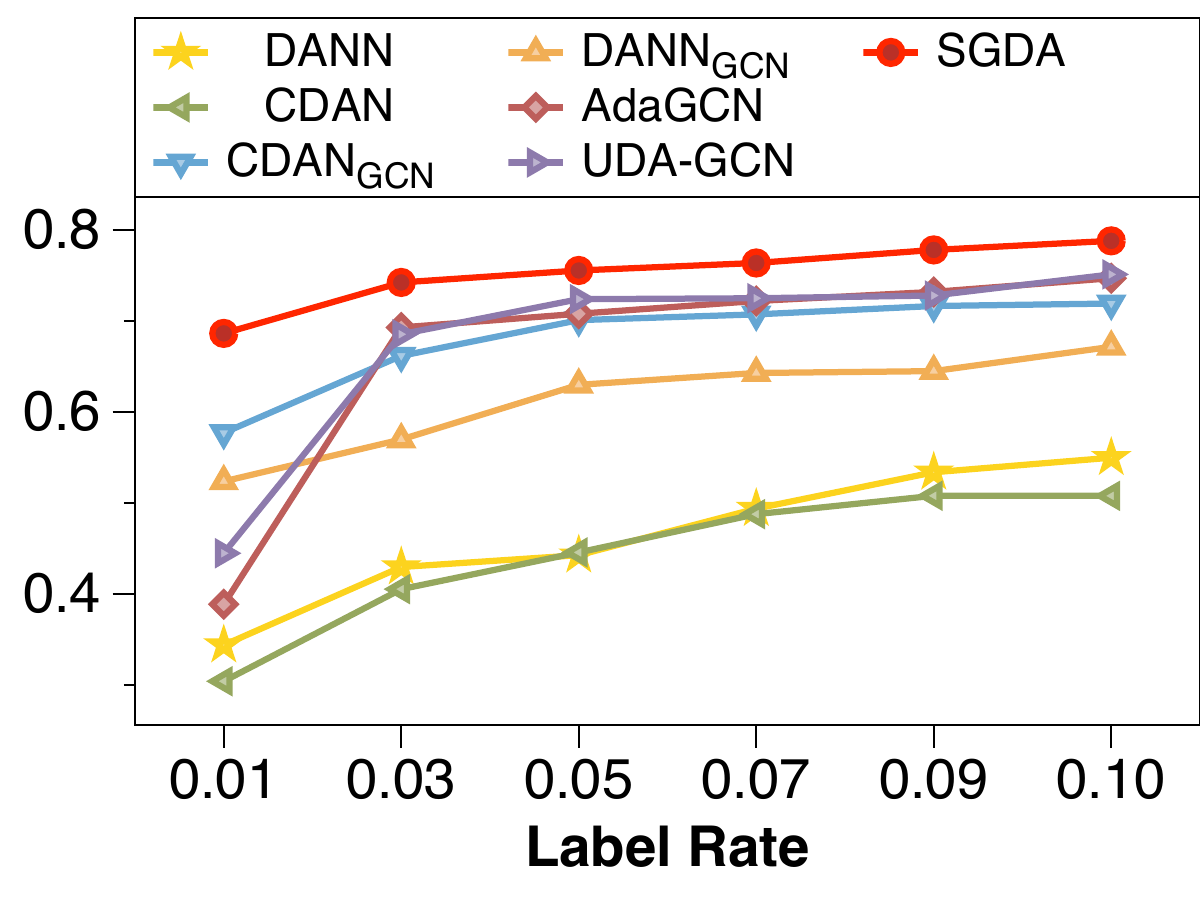}
}
\subfigure{
\includegraphics[width=4.2cm]{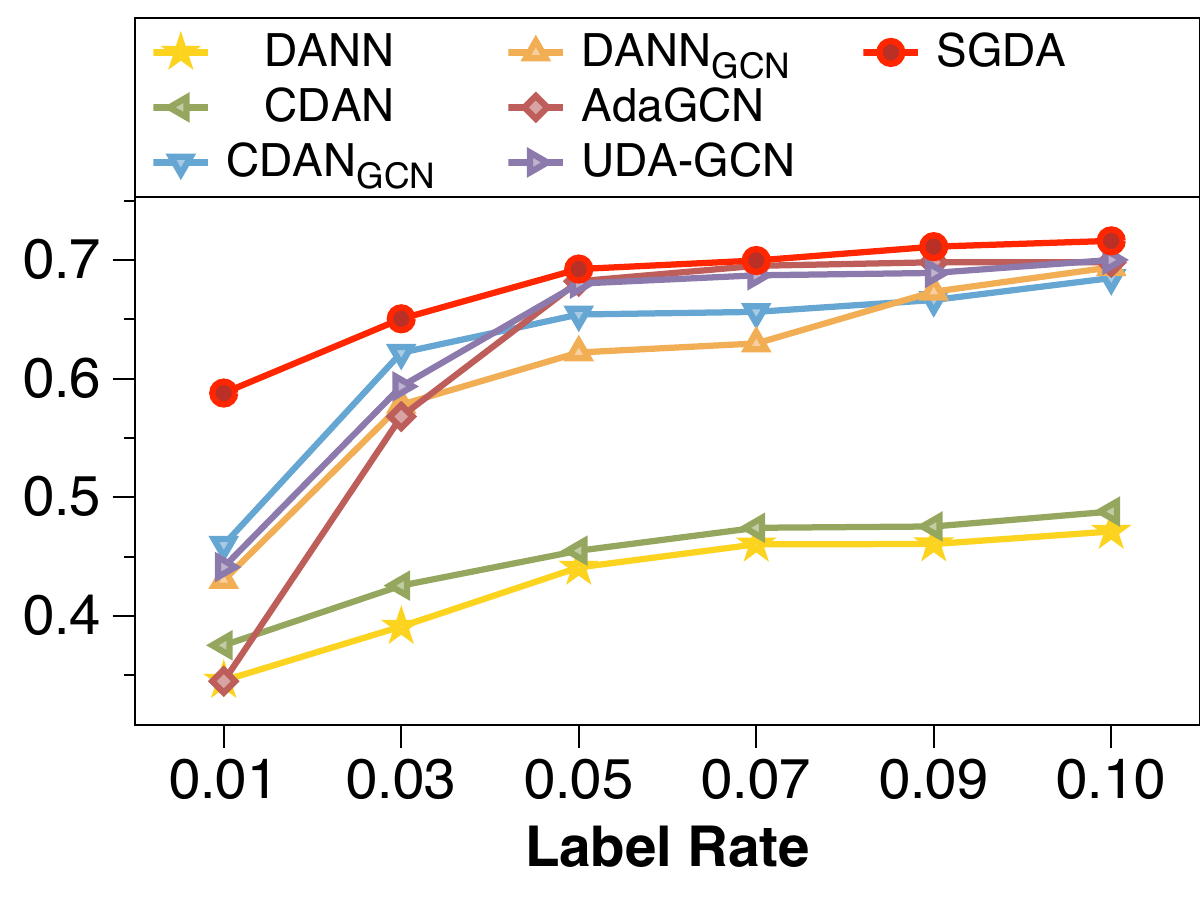}
}
\hspace{-4mm}
    \caption{The model performance with different label rates on the A $\Rightarrow$ C task (left) and the A $\Rightarrow$ D task (right). 
    }
    \label{fig:label_rate}
\end{figure}

\subsection{Hyper-Paramter Experiment}

\textbf{Effect of Label Rate.}
This experiment aims to answer: \textit{Is \method{} robust with different ratios of labeled data on the source graph?}
We evaluate the performance of different methods with the label rate of the source graph as 1\%, 5\%, 7\%, 9\%, and 10\%, respectively. The results are reported in Figure~\ref{fig:label_rate}. The first observation is that \method{} has a remarkable margin compared with other selected baselines, even with only 1\% labeled data in the source graph. This shows the proposed pseudo-labeling can significantly handle the label scarcity problem. 
Also, the GNN-based methods have a great leap compared with NN-based approaches, proving the great potential of utilizing unlabeled graph topology information in improving the model's robustness under limited labels.

\begin{figure}[]
\centering
\subfigure{
\includegraphics[width=4.0cm]{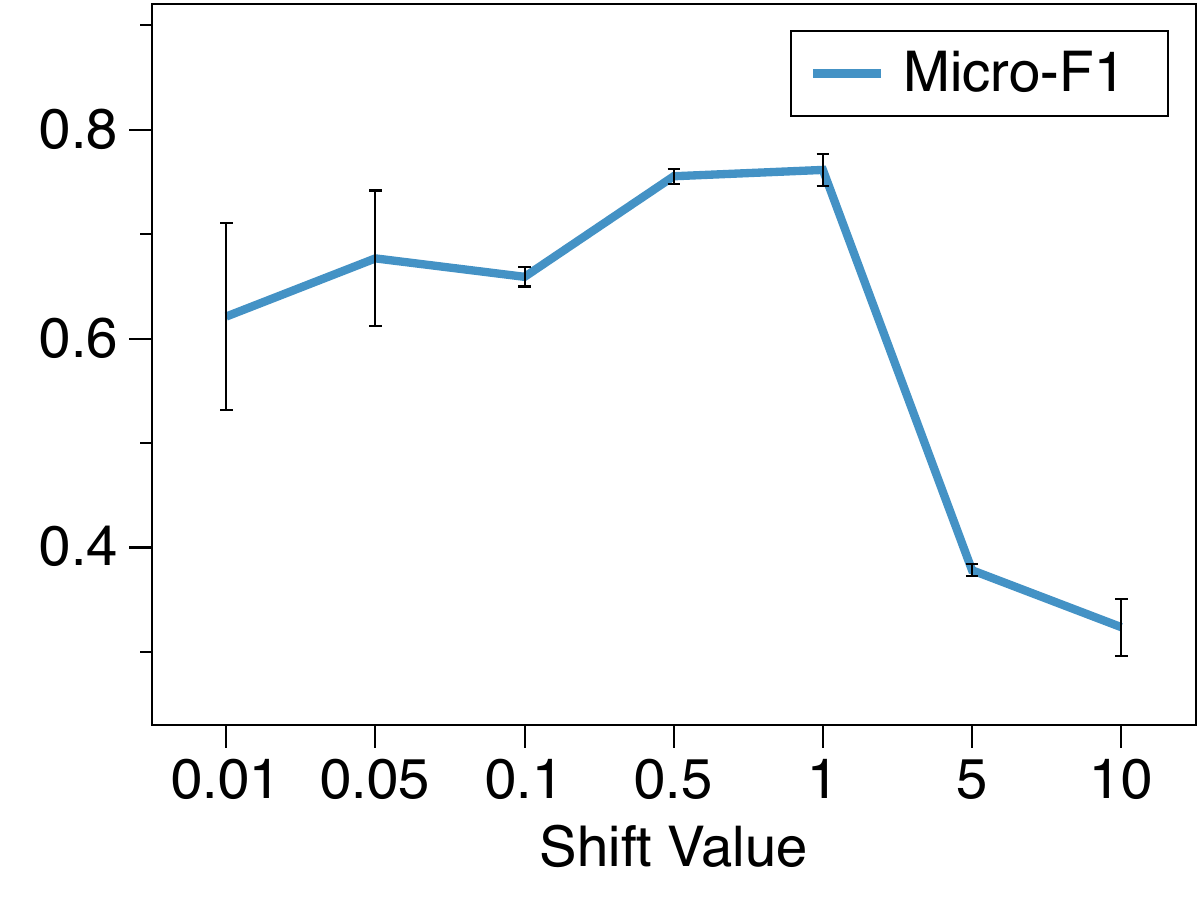}
}
\subfigure{
\includegraphics[width=4.0cm]{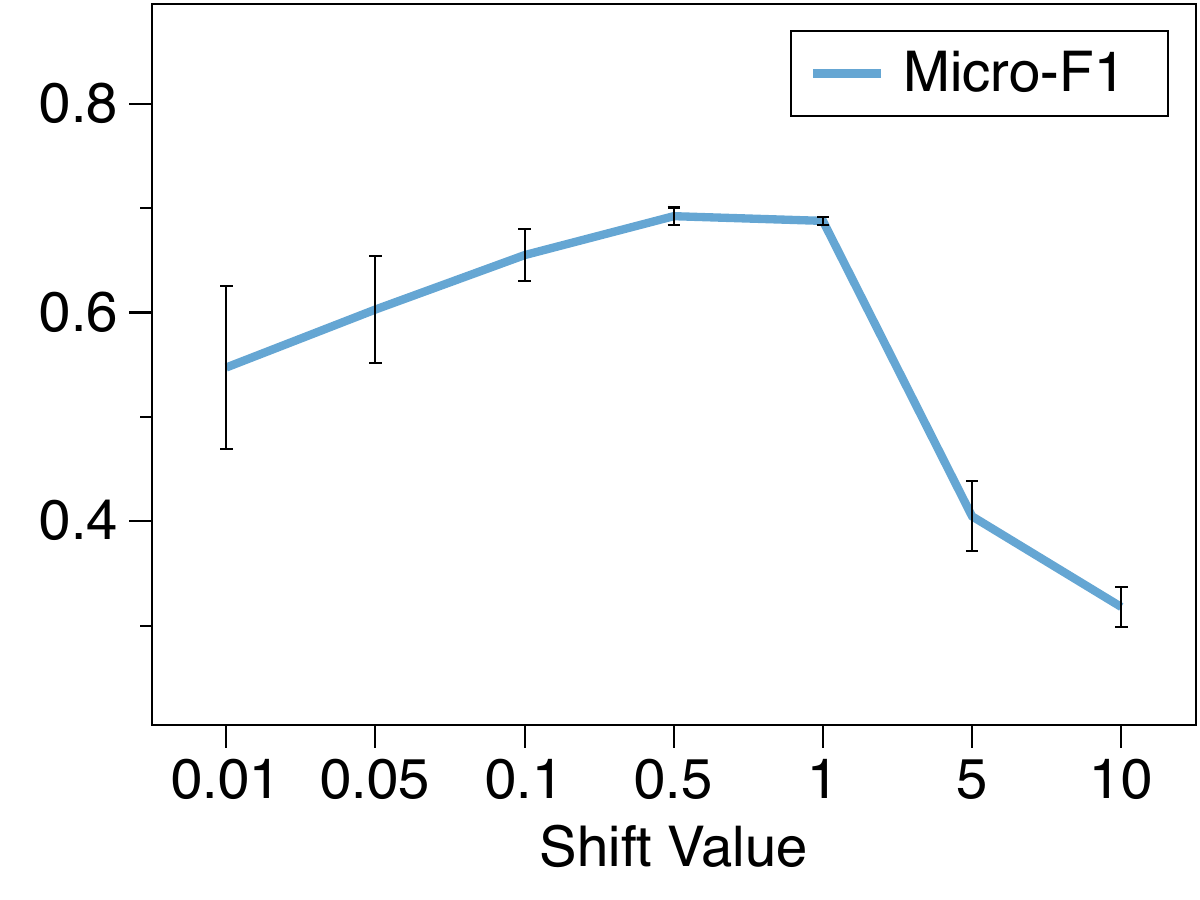}
}
\caption{The model performance with different shift values  on the A $\Rightarrow$ C task (left) and the A $\Rightarrow$ D task (right).}
    \label{fig:per}
\end{figure}

\noindent\textbf{Effect of Shift Value.}
This experiment aims to answer: \textit{How do different shift values affect the performance of \method{}?}
The constraint shift value $\epsilon$ of the shift parameters is significant to control the scale of distribution shifting.
We evaluated \method{} with $\epsilon$ as 0.01, 0.05, 0.1, 0.5, 1. 5, and 10, respectively and report the results in Figure \ref{fig:per}.
We can observe that with low shift values, the model's performance is less robust, showing high standard deviations. Within $\epsilon\in [0.1,1]$, shift parameters have more impact in training, so \method{} can achieve relatively high and stable results. However, when $\epsilon$ is large, the adversarial learning becomes difficult, thus dampening the results.

\section{Conclusion}

This work presents a novel research problem of semi-supervised domain adaptation on graphs.
We propose a method called \method{} that uses shift parameters and adversarial learning 
to achieve model transferring. 
Also, \method{} uses pseudo labels with adaptive posterior scores to alleviate the label scarity. 
Extensive experiments on a variety of publicly available datasets demonstrate the efficacy of \method{}. In future work, we will expand our \method{} to a variety of graph transfer learning tasks including source-free domain adaptation and out-of-domain generalization on graphs.

\clearpage

\section*{Acknowledgements}
This work is supported in part by Foshan HKUST Projects (FSUST21-FYTRI01A, FSUST21-FYTRI02A) and the Natural Science Foundation of China under Grant No. 61836013.

\section*{Contribution Statement}
In this work, Ziyue Qiao, Xiao Luo, and, Meng Xiao contributed equally.
Specifically, Ziyue Qiao and Xiao Luo contributed to the problem formulation, methodology, model implementation, and paper writing. Meng Xiao contributed to the conduction, visualization, and writing of experiments. 
Hao Dong assisted in the paper writing. Yuanchun Zhou and Hui Xiong provided valuable feedback on the paper drafts. All authors reviewed and approved the final manuscript.

\bibliographystyle{named}
\bibliography{sample-base}

\end{document}